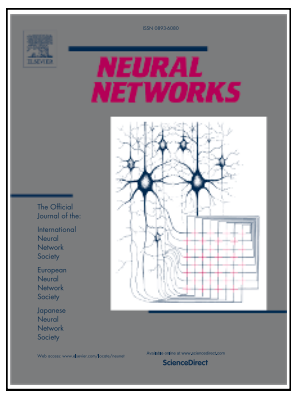

Full Length Article

# Key-value pair-free continual learner via task-specific prompt-prototype

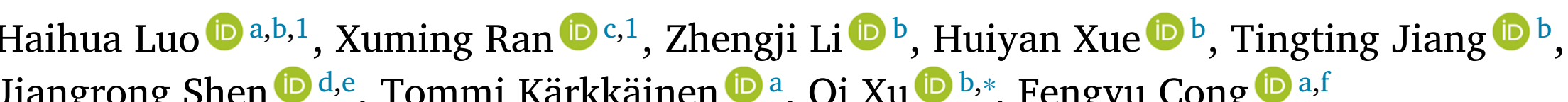

Haihua Luo [a,b,1], Xuming Ran [c,1], Zhengji Li [b], Huiyan Xue [b], Tingting Jiang [b], Jiangrong Shen [d,e], Tommi Kärkkäinen [a], Qi Xu [b,*], Fengyu Cong [a,f]

[a] *University of Jyväskylä, Faculty of Information and Technology, Finland*
[b] *Dalian University of Technology, School of Computer Science and Technology, China*
[c] *National University of Singapore, Singapore*
[d] *Zhejiang University, Lab of Brain-Machine Intelligence, China*
[e] *Xi'an Jiaotong University State Key, School of Computer Science and Technology, China*
[f] *Dalian University of Technology, School of Biomedical Engineering, China*

ABSTRACT

Continual learning aims to enable models to acquire new knowledge while retaining previously learned information. Prompt-based methods have shown remarkable performance in this domain; however, they typically rely on key-value pairing, which can introduce inter-task interference and hinder scalability. To overcome these limitations, we propose a novel approach employing task-specific **Pro**mpt-**P**rototype (**ProP**), thereby eliminating the need for key-value pairs. In our method, task-specific prompts facilitate more effective feature learning for the current task, while corresponding prototypes capture the representative features of the input. During inference, predictions are generated by binding each task-specific prompt with its associated prototype. Additionally, we introduce regularization constraints during prompt initialization to penalize excessively large values, thereby enhancing stability. Experiments on several widely used datasets demonstrate the effectiveness of the proposed method. In contrast to mainstream prompt-based approaches, our framework removes the dependency on key-value pairs, offering a fresh perspective for future continual learning research.

## 1. Introduction

In recent years, neural network models based on deep learning have made significant advances. However, in real-world scenarios, models are continually confronted with new tasks. When learning new tasks, models often forget the knowledge acquired from previous ones, which is known as Catastrophic Forgetting (Goodfellow et al., 2013). We aspire for models to possess the ability to learn continuously, akin to humans, which means learning new tasks without forgetting the knowledge of old ones. To this end, researchers propose various continual learning methods (Lange et al., 2021; Masana et al., 2022). However, traditional continual learning approaches (Li & Hoiem, 2017; Rebuffi et al., 2017; Yan et al., 2021) train models from scratch, making it challenging to learn high-quality features from data samples. Pre-trained model(PTM) (Dosovitskiy, 2020; He et al., 2022; Radford et al., 2021), having been thoroughly trained on large-scale datasets, possesses outstanding feature extraction capabilities. Introducing pre-trained models into continual learning can significantly enhance the feature extraction performance of the model. In recent years, continual learning based on pre-trained models has become increasingly widespread. Nevertheless, due to the domain gap (You et al., 2020; Zhou et al., 2022b) between pre-training datasets and downstream datasets in continual learning, pre-trained models require finetuning to better adapt to downstream tasks.

Among various techniques for finetuning pre-trained models in continual learning, prompt-based methods (Liu et al., 2023; Ran et al., 2024; Zhou et al., 2022c) demonstrate exceptional performance. A prompt is a set of learnable vectors whose dimensions match the hidden features. During training, the weights of the pre-trained model are frozen, and the prompt is directly concatenated with the input embedding to form a new composite embedding. By updating the prompt, the model is able to learn task-specific features pertinent to downstream tasks. L2P (Wang et al., 2022d) is the first to introduce prompt into continual learning by proposing a shared prompt pool for all tasks and using key-value pairing to query the prompt that matches the task. Building upon this, Dual Prompt (Wang et al., 2022c) incorporates a global prompt to capture information shared across tasks. Coda-Prompt (Smith et al., 2023) replaces the prompt pool with a prompt component pool and utilizes a weighted sum of prompt components for initialization. However, these methods all depend on key-value pairing during inference, as shown in Fig. 1(a), which requires matching the current data sample to its corresponding prompt from the prompt pool. The prompt pool contains prompts for all tasks, therefore the key-value pairing inevitably introduces inter-task interference (Kurniawan et al., 2024; Wang et al., 2025), as prompts from different tasks can affect the matching process. For instance, when the input sample is a Persian cat, the similarity in features between Persian cats and tabby cats may lead to an incorrect match with the key corresponding to tabby cat during key–value pairing. Moreover, as the number of tasks increases, the number of key-value pairs grows correspondingly. The model must rapidly retrieve matching keys from a vast number of options, increasing computational and memory overhead (Yang et al., 2024).

* Corresponding author.
*E-mail addresses:* haihua.h.luo@jyu.fi (H. Luo), xuqi@dlut.edu.cn (Q. Xu).

[1] Equal contribution.

To address these issues, this paper proposes ProP, which eliminates the dependency on key-value pairs by binding prompt with prototype, as shown in Fig. 1(b). During training, ProP initializes task-specific prompt instead of selecting from a shared prompt pool and obtains the corresponding task-specific prototype after training. Prototype (Snell et al., 2017) is the representative feature of the current class. In this work, we use the mean of the class's features as its prototype. For data belonging to the same class, their features should exhibit the highest similarity with that prototype of the class. By computing the similarity between features and various prototypes, we can determine the probability that the data belongs to each class. Furthermore, during prompt initialization, random initialization may cause the prompt to attain extreme values, leading to significant deviations in feature learning. Therefore, we introduce a regularization loss in the prompt initialization to penalize excessively large initial values, aiming to obtain stable and more generalizable prompts. As shown in Fig. 2, ProP archives the best performance compared to other CIL methods without the requirement of exemplars. The contributions of this paper are summarized as follows:

- We propose ProP, a simple framework that outperforms existing Class-Incremental Learning(CIL) methods across a wide range of datasets. ProP eliminates the reliance on key-value pairs in previous mainstream prompt-based methods, avoiding the inter-task information interference and scalability limitations caused by key-value pairs.
- We introduce task-specific prompt-prototype by binding task-specific prompt with prototype sharing same task id, thereby integrating task-specific information into the model learning process without additional retrieval steps.
- We incorporate a regularization loss in prompt initialization to penalize excessively large initial values, rendering prompt initialization more stable and generalizable.

## 2. Related work

### 2.1. Conventional continual learning

Continual learning methods aim to enable models to retain previously acquired knowledge while learning new tasks, thereby alleviating the problem of catastrophic forgetting. Conventional continual learning methods can be broadly classified into three categories: replay-based, regularization-based, and network structure-based. Replay-based methods (Rebuffi et al., 2017; Wang et al., 2022a; Yan et al., 2021; Zhou et al., 2022a)help models preserve prior knowledge when learning new tasks by storing and replaying examples of old classes. However, the storage of examples from old classes raises concerns about the leakage of private information. Additionally, the size of the buffer may increase with the number of tasks, which leads to escalating memory requirements. Regularization-based methods (Kirkpatrick et al., 2017; Li & Hoiem, 2017; Zenke et al., 2017) impose constraints on the parameters update of model to maintain the stability of key features across different tasks, effectively preserving knowledge of previous tasks. However, this also limits the ability of model to learn new classes. Network structure-based methods (Chen & Chang, 2023; Douillard et al., 2022; Hu et al., 2023) dynamically adjust the model architecture to avoid interference from parameter updates on previous tasks. Nevertheless, as the number of tasks increases, the model size can grow rapidly, reducing training and inference efficiency and limiting information sharing between tasks.

### 2.2. Pre-trained model for continual learning

Pre-trained models (Dosovitskiy, 2020; He et al., 2022; Radford et al., 2021) are thoroughly trained on large-scale datasets and possess exceptional feature extraction capabilities. For example, the Vision Transformer (ViT) (Dosovitskiy, 2020). By integrating pre-trained models into continual learning, the performance of model has been significantly improved (Mcdonnell et al., 2024; Zhou et al., 2024b; Xue et al., 2025). However, there is a domain gap between the pre-trained datasets and the continual learning downstream datasets, making it challenging for pre-trained models to effectively learn robust feature representations tailored to downstream data when applied directly. Consequently, finetuning of pre-trained models is necessary to adapt them to specific downstream tasks.

In recent years, a multitude of finetune techniques have emerged, including Adapter (Houlsby et al., 2019), Prompt (Jia et al., 2022), and Low-Rank Adaptation (LoRA) (Hu et al., 2021). These approaches typically involve freezing the pre-trained weights while finetuning only task-specific components during the training process. This strategy enables models to adapt efficiently to new tasks with minimal parameter updates. By frozen the weights of pre-trained models, model maximally retains the exceptional feature extraction capacity of pre-trained models. Such methods demonstrate how pre-trained models can serve as stable backbones in continual learning, allowing for the integration of new tasks with limited adjustments and reduced computational costs.

### 2.3. Prompt for continual learning

Prompt-based methods (Wang et al., 2024, 2022b; Zhou et al., 2024a) introduce learnable prompt vectors to guide pre-trained models in task-specific learning while minimizing parameter updates, garnering significant attention in the field of continual learning. L2P (Wang et al., 2022d) is the first to incorporate prompt into continual learning by establishing a shared prompt pool for all tasks to learn task-specific features and employing key-value pairs to associate the appropriate prompts. Building on L2P, Dual Prompt (Wang et al., 2022c) introduces a global prompt to capture information shared across tasks. Coda-Prompt (Smith et al., 2023) replaces the traditional prompt pool with a prompt component pool and initializes prompts through a weighted sum of prompt components.

However, these approaches rely on key-value pairs to match prompts corresponding to the current data samples from the prompt pool during inference. Since the prompt pool stores prompts of all tasks, the key-value pairing process is inevitably susceptible to interference from other prompts, meaning that key-value pairs can be affected by inter-task information (Kurniawan et al., 2024). For instance, when the input sample is a Persian cat, the similarity in features between Persian cats and tabby cats may lead to an incorrect match with the key corresponding to tabby cat during key–value pairing. Additionally, as the number of tasks increases, the number of key-value pairs also grows. The model must rapidly retrieve matching keys from a vast number of keys, leading to increased demands on memory and computational resources (Yang et al., 2024).

## 3. Method

In this chapter, we first introduce prompt-based continual learning in Section 3.1, including the description of continual learning, pre-trained models, and prompt methods. Then, we present the task-specific prompt-prototype and describe its optimization process in Section 3.2. We further provide pseudo code of training and test stage in Section 3.3 and finally, we analyze the computational complexity of each continuous learning method in Section 3.4.

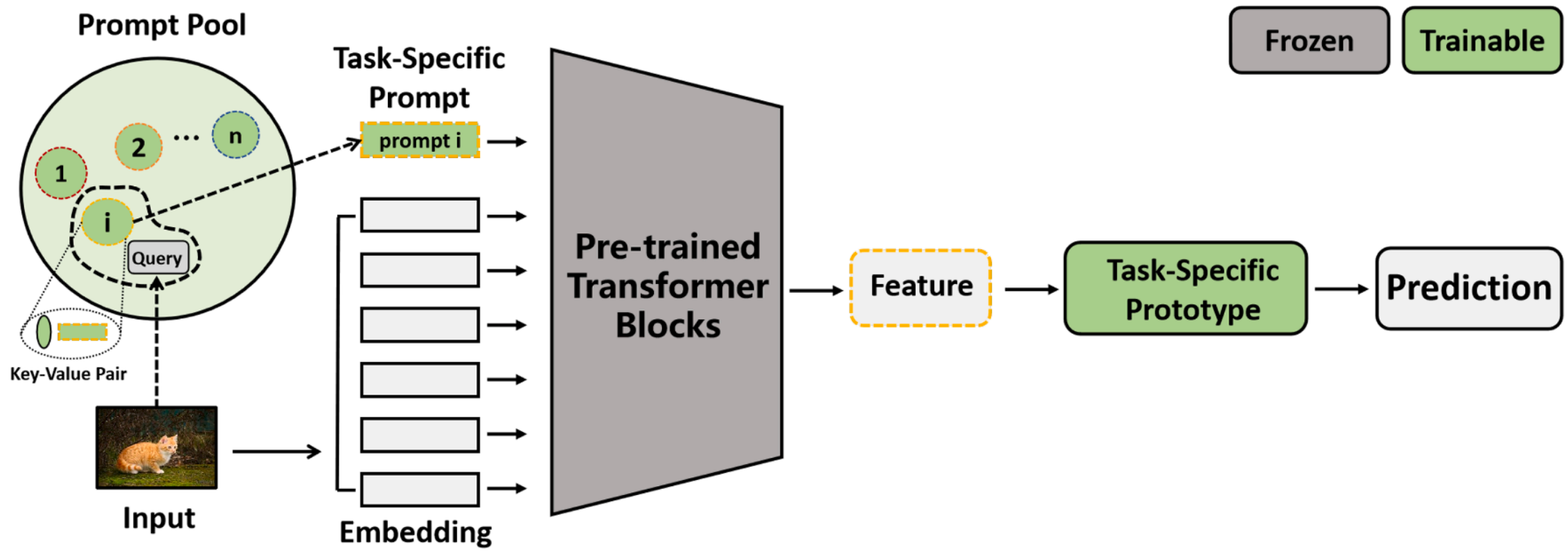

(a) Framework of other prompt-based methods

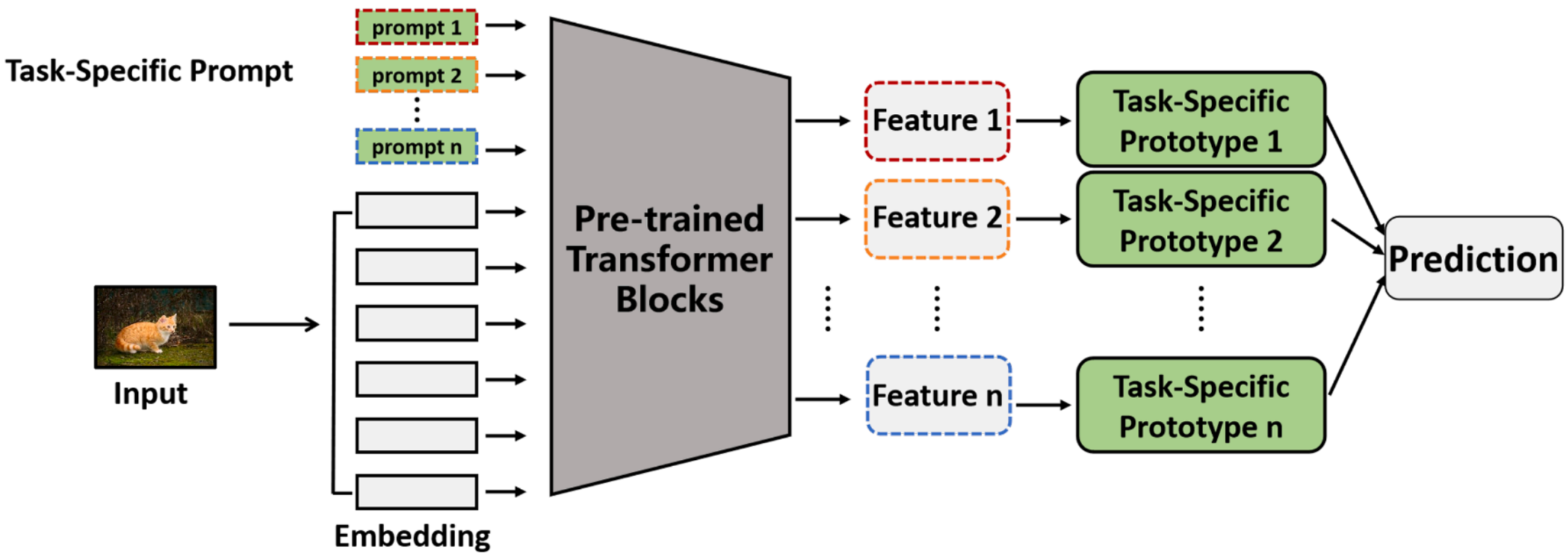

(b) Framework of ProP

**Fig. 1.** Comparison of frameworks between mainstream prompt-based methods and ProP. Color green means trainable at training, while gray means frozen the components. (a) illustrates the inference process of mainstream prompt-based methods, which relies on key-value pairs to determine the prompt corresponding to the input. (b) shows the inference process of ProP, which binds task-specific prompt with prototype sharing same task id without the need for key-value pairs.

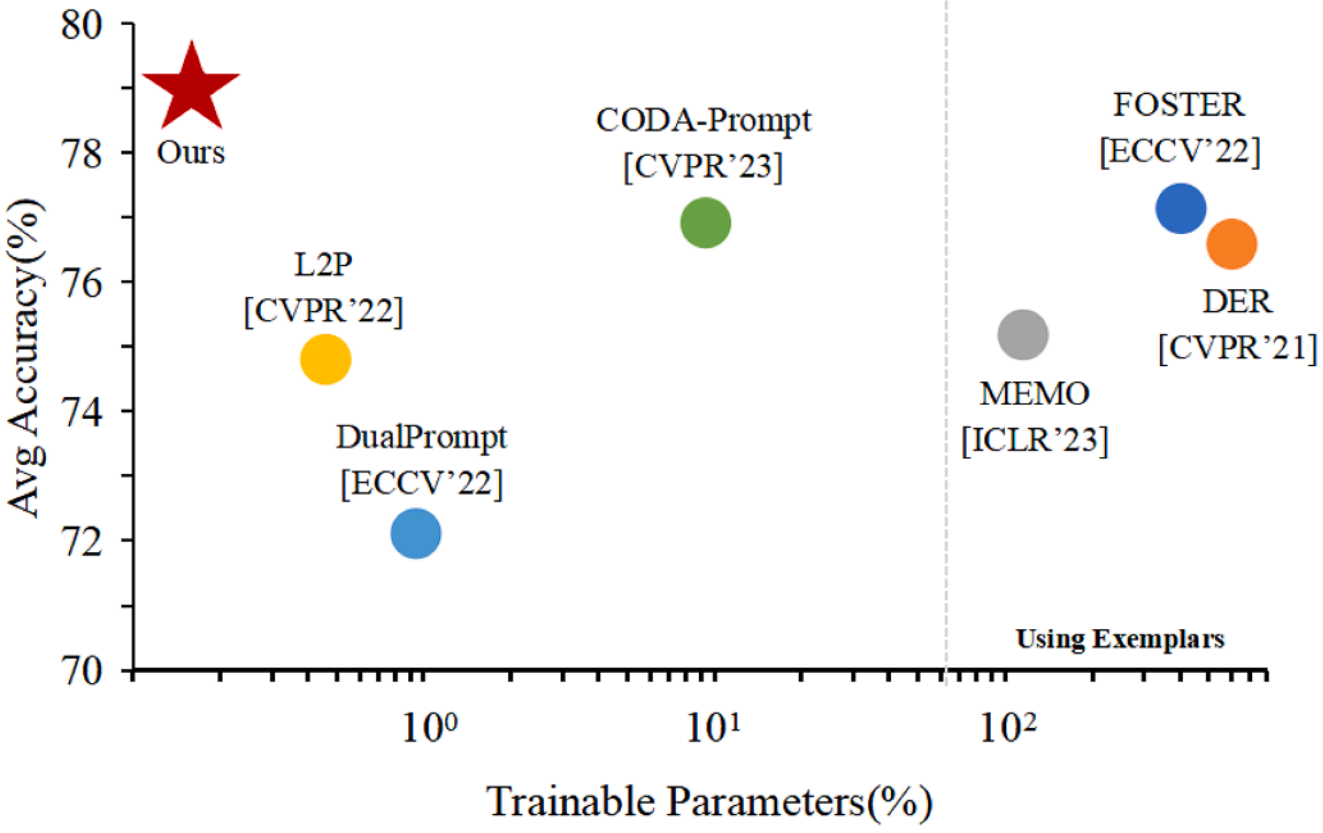

**Fig. 2.** Comparison of trainable parameters and average accuracy between ProP (Ours) and other CIL methods. "Using Exemplars" indicates storing samples from previous tasks for rehearsal. All experiments are conducted on ImageNet-R Init100 Inc50, using the same pre-trained model as the backbone.

### 3.1. Preliminaries

**Continual Learning.** Continual learning faces a sequence of tasks $\mathcal{D}_1, \ldots, \mathcal{D}_T$. The $t$th task is defined as $\mathcal{D}_t = \{(\boldsymbol{x}_i^t, \boldsymbol{y}_i^t)\}_{i=1}^{m_t}$, where $\mathcal{D}_t$ contains $m_t$ samples $\boldsymbol{x}_i^t$ and their corresponding labels $\boldsymbol{y}_i^t$. It is worth noting that in this paper, we focus on CIL without rehearsal, which means the data $\mathcal{D}_1, \mathcal{D}_2, \ldots, \mathcal{D}_{t-1}$ are unavailable during the learning of the current task $\mathcal{D}_t$. Moreover, unlike Task-Incremental Learning (TIL) and Domain-Incremental Learning (DIL), in CIL the task id is unknown during testing, and the model needs to directly infer the class to which the data belongs, making it more closely aligned with real-world scenarios (Ven et al., 2022).

**Prompt for Continual Learning.** Prompt (Jia et al., 2022) is a trainable vector used to finetune pre-trained model, such as ViT (Dosovitskiy, 2020), with minimal parameter updates. Here, we denote pre-trained model as $f(\theta)$. The prompt $\boldsymbol{p} \in \mathbb{R}^{L_p \times D}$ shares the same feature dimension $D$ as the hidden features $\boldsymbol{h} \in \mathbb{R}^{L_h \times D}$ and can be directly concatenated with $\boldsymbol{h}$, where $L_h$ is the sequence length of the hidden features, $L_p$ is the length of the prompt. As the number of training tasks increases, each task $\mathcal{D}_t$ generates a corresponding prompt $\boldsymbol{p}_t \in \mathbb{R}^{L_p \times D}$, ultimately forming a prompt pool $\boldsymbol{P} \in \mathbb{R}^{T \times L_p \times D}$.

During training, the parameters of the prompt are updated to finetune the pre-trained model for continual learning tasks, while the parameters of pre-trained model $\theta$ remain frozen to preserve its feature extraction capabilities. Specifically, for a multi-head self-attention layer (MSA) that contains query $\boldsymbol{h}_Q$, key $\boldsymbol{h}_K$, and value $\boldsymbol{h}_V$, the output of the

**Table 1**
Average and last performance comparison on seven datasets with **ViT-B/16-IN21K** as the backbone. "Init5 Inc5" means 5 classes in first task and 5 classes in increment tasks. 'IN-R/A' stands for 'ImageNet-R/A,' 'ObjNet' stands for 'ObjectNet,' and 'OmniBench' stands for 'OmniBenchmark.' The best performance is shown in bold.

| Method | CIFAR Init5 Inc5 | | CUB Init10 Inc10 | | IN-R Init5 Inc5 | | IN-A Init20 Inc20 | | ObjNet Init10 Inc10 | | OmniBench Init30 Inc30 | | VTAB Init10 Inc10 | |
|---|---|---|---|---|---|---|---|---|---|---|---|---|---|---|
| | Avg | Last | Avg | Last | Avg | Last | Avg | Last | Avg | Last | Avg | Last | Avg | Last |
| Finetune | 38.90 | 20.17 | 26.08 | 13.96 | 21.61 | 10.79 | 24.28 | 14.51 | 19.14 | 8.73 | 23.61 | 10.57 | 34.95 | 21.25 |
| Finetune Adapter | 60.51 | 49.32 | 66.84 | 52.99 | 47.59 | 40.28 | 45.41 | 41.10 | 50.22 | 35.95 | 62.32 | 50.53 | 48.91 | 45.21 |
| LwF | 46.29 | 41.07 | 48.97 | 32.03 | 39.93 | 26.47 | 37.75 | 26.84 | 33.01 | 20.65 | 47.14 | 33.95 | 40.48 | 27.54 |
| SDC | 68.21 | 63.05 | 70.62 | 66.37 | 52.17 | 49.20 | 29.11 | 26.63 | 39.04 | 29.06 | 60.94 | 50.28 | 45.06 | 22.50 |
| L2P | 85.94 | 79.93 | 67.05 | 56.25 | 66.53 | 59.22 | 49.39 | 41.71 | 63.78 | 52.19 | 73.36 | 64.69 | 77.11 | 77.10 |
| DualPrompt | 87.87 | 81.15 | 77.47 | 66.54 | 63.31 | 55.22 | 53.71 | 41.67 | 59.27 | 49.33 | 73.92 | 65.52 | 83.36 | 81.23 |
| CODA-Prompt | 89.11 | 81.96 | 84.00 | 73.37 | 64.42 | 55.08 | 53.54 | 42.73 | 66.07 | 53.29 | 77.03 | 68.09 | 83.90 | 83.02 |
| APER(VPT-Shallow) | 90.43 | 84.57 | **92.02** | **86.51** | 66.63 | 58.32 | 58.39 | 47.20 | 64.54 | 52.53 | 79.63 | 73.68 | 87.15 | 85.36 |
| APER(VPT-Deep) | 88.46 | 82.17 | 91.02 | 84.99 | 68.79 | 60.48 | 58.48 | 48.52 | 67.83 | 54.65 | 81.05 | 74.47 | 86.59 | 83.06 |
| ProP(Ours) | **91.84** | **85.99** | 91.20 | 84.82 | **77.50** | **68.78** | **66.57** | **55.96** | **72.34** | **60.19** | **82.01** | **75.29** | **94.77** | **93.91** |

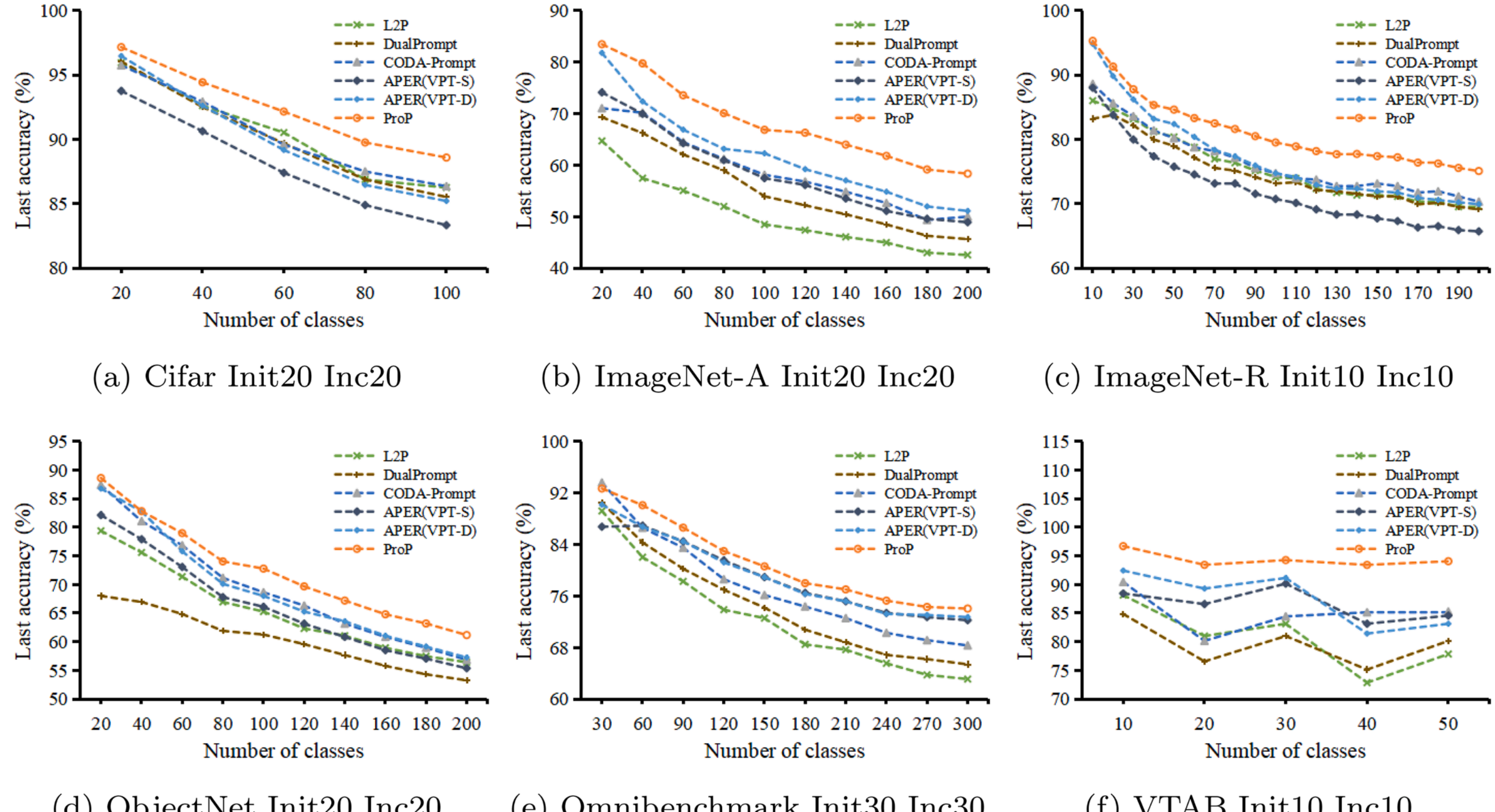

**Fig. 3.** Last accuracy under different incremental settings, with the number of classes in the initial task kept the same as in the incremental tasks. All experiments are conducted using **ViT-B/16-IN1K** as the backbone.

MSA is computed as:

$$\mathrm{MSA}(\boldsymbol{h}_Q, \boldsymbol{h}_K, \boldsymbol{h}_V) = \mathrm{Concat}(h_1, \ldots, h_s)W^O, \tag{1}$$

$$h_i = \mathrm{Attention}(\boldsymbol{h}_Q W_i^Q, \boldsymbol{h}_K W_i^K, \boldsymbol{h}_V W_i^V) \tag{2}$$

where $W^O$, $W_i^Q$, $W_i^K$, and $W_i^V$ are projection matrices, and $s$ is the number of attention heads. For pre-trained model like ViT, the queries, keys, and values are the same, i.e., $\boldsymbol{h} = \boldsymbol{h}_Q = \boldsymbol{h}_K = \boldsymbol{h}_V$. To finetune the pre-trained model, the input $\boldsymbol{h}$ is concatenated with the prompt $\boldsymbol{p}$:

$$f_{\text{prompt}} = \mathrm{MSA}([\boldsymbol{h}_Q; \boldsymbol{p}], [\boldsymbol{h}_K; \boldsymbol{p}], [\boldsymbol{h}_V; \boldsymbol{p}]) \tag{3}$$

where $[,;,]$ denotes concatenation along the sequence dimension. The output of the MSA maintains the same dimension as the original input $\boldsymbol{h} \in \mathbb{R}^{L_h \times D}$. Accordingly, we denote the finetuned pre-trained model as $f(\theta, \boldsymbol{p})$.

**Key-Value Pair in Prompt-based Continual Learning.** Prompt-based continual learning methods demonstrate excellent performance, however, mainstream prompt-based approaches often rely on key-value pairing to search for task-specific prompt $\boldsymbol{p}$ within prompt pool $\boldsymbol{P}$. For the $t$th task, the knowledge of task $D_t$ is learned during training based on prompt $\boldsymbol{p}_t$. However, during inference when the task id is unknown, mainstream prompt-based methods typically need to compute a query by pre-trained model, usually represented as:

$$\boldsymbol{q} = f(\theta, \boldsymbol{x}) \tag{4}$$

where $\boldsymbol{q} \in \mathbb{R}^{L_h \times D}$ is the query of input $\boldsymbol{x}$. The corresponding prompt $\boldsymbol{p}_i$ of input $\boldsymbol{x}$ is chosen by matching the query $\boldsymbol{q}$ to key $\boldsymbol{k}_i \in \mathbb{R}^D$ with the highest similarity in the key pool $\boldsymbol{E} \in \mathbb{R}^{T \times D}$:

$$\boldsymbol{p}_i = \gamma(\boldsymbol{q}, \boldsymbol{k}_i) \tag{5}$$

Since the prompt pool $\boldsymbol{P}$ contains all prompts $\boldsymbol{p}_t (t = 1, \ldots, T)$, such as Persian cat and tabby cat, the model may retrieve an incorrect task ID, leading to erroneous predictions. Furthermore, as the number of tasks $T$ increases, the number of key-value pairs may grows correspondingly.

The model need to rapidly retrieve matching keys from a vast number of options, which increases both computational and memory overhead.
**Prototype.** The prototype of a class contains the representative features of that class. In this paper, we follow Snell et al. (2017), setting the mean of the features of class as its prototype. Specifically, for the $t$th task $\mathcal{D}_t = \{\boldsymbol{x}_i^t, \ldots, \boldsymbol{x}_i^t\}_{i=1}^{n_t}$, after the model has been trained, we compute the prototype $\boldsymbol{c}_m \in \mathbb{R}^D$ of class $m$ in $t$th task $\mathcal{D}_t$:

$$\boldsymbol{c}_m = \frac{1}{M}\sum_{j=1}^{|\mathcal{D}_t|} I(y_j = m) f(\theta, \boldsymbol{p}, \boldsymbol{x}) \tag{6}$$

where $M = \sum_{j=1}^{||\mathcal{D}_t||} I(y_j = m)$, $I(\cdot)$ is the indicator function. Since $\boldsymbol{c}_m$ represents the average features of class $m$, this implies that images belonging to class $m$ should exhibit the greatest similarity to $\boldsymbol{c}_m$ among all prototypes.

### 3.2. Proposed approach

**Task-Specific Prompt-Prototype.** For each $t$th task, we generate a corresponding prompt $\boldsymbol{p}_t$ to better capture and learn the features specific to that task $\mathcal{D}_t$. During training, we finetune the pre-trained model based on Eq. (3) to obtain enhanced feature representations of the input $\boldsymbol{x}$. Concurrently, using Eq. (6), we derive the prototype $\boldsymbol{c}_{t,\boldsymbol{p}_t}$ for task $t$ under prompt $\boldsymbol{p}_t$. This prototype $\boldsymbol{c}_{t,\boldsymbol{p}_t} \in \mathbb{R}^D$ represents the characteristic features of classes in task $t$ learned through prompt $\boldsymbol{p}_t$.

$$\boldsymbol{c}_{t,\boldsymbol{p}_t} = \frac{1}{M}\sum_{j=1}^{|\mathcal{D}_t|} I(y_j = m) f(\theta, \boldsymbol{p}_t, \boldsymbol{x}) \tag{7}$$

To maximally preserve the feature extraction capability of pre-trained model, we also compute the prototype $\boldsymbol{c}_{t,\theta} \in \mathbb{R}^D$ corresponding to classes in task $\mathcal{D}_t$ using the pre-trained model.

$$\boldsymbol{c}_{t,\theta} = \frac{1}{M}\sum_{j=1}^{|\mathcal{D}_t|} I(y_j = m) f(\theta, \boldsymbol{x}) \tag{8}$$

We form the task-specific prototype $\boldsymbol{C}_t \in \mathbb{R}^{2D}$ for classes in task $\mathcal{D}_t$ by concatenating $\boldsymbol{c}_{t,\boldsymbol{p}_t}$ and $\boldsymbol{c}_{t,\theta}$:

$$\boldsymbol{C}_t = [\boldsymbol{c}_{t,\boldsymbol{p}_t}; \boldsymbol{c}_{t,\theta}] \tag{9}$$

During task-agnostic inference, the input $\boldsymbol{x}$ is processed through each prompt $\boldsymbol{p}_i (i = 1, \ldots, t)$ and the pre-trained model to obtain the corresponding features $\boldsymbol{h}_{\boldsymbol{p}_i} \in \mathbb{R}^D$ and $\boldsymbol{h}_\theta \in \mathbb{R}^D$: $\boldsymbol{h}_{\boldsymbol{p}_i} = f(\theta, \boldsymbol{p}_i, \boldsymbol{x})$ and $\boldsymbol{h}_\theta = f(\theta, \boldsymbol{x})$. The feature representation $\boldsymbol{H}_{\boldsymbol{p}_i} \in \mathbb{R}^{2D}$ of input $\boldsymbol{x}$ under prompt $\boldsymbol{p}_i$ is formed by concatenating them:

$$\boldsymbol{H}_{\boldsymbol{p}_i} = [\boldsymbol{h}_{\boldsymbol{p}_i}; \boldsymbol{h}_\theta] \tag{10}$$

For inference, we generate features $\boldsymbol{H}_{\boldsymbol{p}_i}$ of input $\boldsymbol{x}$, which are compared against their respective prototypes $\boldsymbol{C}_i$ by calculating their similarity. Instead of explicitly determining the task id of the input by key-value pairing, we bind prompt with prototype sharing the same task id and directly compute the similarity between each task-specific feature $\boldsymbol{H}_{\boldsymbol{p}_i}$ and its corresponding prototype $\boldsymbol{C}_i$.

$$\phi(\boldsymbol{x}) = \left(\frac{\mathbf{H}_{\mathbf{p_i}}}{\|\mathbf{H}_{\mathbf{p_i}}\|_2}\right)^\top \left(\frac{\boldsymbol{C}_i}{\|\boldsymbol{C}_i\|_2}\right) \tag{11}$$

Here $\phi(\boldsymbol{x})$ is the classifier function of model. Compared to other prompts, the prompt $\boldsymbol{p}$ corresponding to the input $\boldsymbol{x}$ has been specifically trained for that task, enabling it to learn a robust feature representation for $\boldsymbol{x}$. Consequently, the similarity with the corresponding task-specific prototype should yield the highest score, effectively identifying the correct class without requiring task id information.

In ProP, each task-specific prompt $\boldsymbol{p}_t$ learns to modulate the frozen pre-trained backbone to extract features optimal for task $D_t$. The corresponding prototype $\boldsymbol{C}_t$(Eq. (9)) serves as a stable anchor in the feature space, representing the central tendency of its class. Binding the prompt $\boldsymbol{p}_t$ directly to the prototype $\boldsymbol{C}_t$ effectively creates a dedicated and compact representation subspace for task $t$. During inference, the similarity calculation (Eq. (11)) between the input feature $\boldsymbol{h}_{\boldsymbol{p}_i}$ and all prototypes $\{\boldsymbol{C}_1, \boldsymbol{C}_2, \ldots, \boldsymbol{C}_T\}$ functions as a form of nearest class mean classification within these stable subspaces. This design mitigates inter-task interference in two ways: The frozen pre-trained model provides a stable feature foundation; The task-specific prototypes anchor the decision boundaries, making them more resilient to distortion caused by the learning of new tasks. While in key-value methods, the retrieval step can be contaminated by similar features from other tasks, leading to erroneous prompt selection and subsequent feature distortion.

**Optimization of Prompt-Prototype for Finetuning.** During the training process, both the prompts and the classifier are trainable. For each task, a fully connected (FC) layer $\phi(\boldsymbol{x})$ is randomly initialized, and the prediction of model is given by $\phi(f(\theta, \boldsymbol{p}, \boldsymbol{x}))$. In this study, we first optimize the prompt parameters using the cross-entropy loss (Mao et al., 2023) function between the predicted values and the data labels:

$$\mathcal{L}_{CE} = -\frac{1}{N}\sum_{i=1}^{N}\sum_{j=1}^{M} \boldsymbol{y}_{i,j} \log\left(\phi(f(\theta, \boldsymbol{p}, \boldsymbol{x}_i))_j\right) \tag{12}$$

Here, $\boldsymbol{y}_{i,j}$ is the value of the true label for the $i$th sample at the $j$th class position, $N$ is the batchsize and $\phi(f(\theta, \boldsymbol{p}, x_i))_j$ is the predicted probability of the $i$th sample for the $j$th class by the model.

Furthermore, during the prompt initialization process, random initialization can produce extreme values that may mislead the feature learning. To address this, we incorporate an L2 loss $\mathcal{L}_{L2}$ (Bühlmann & Yu, 2003) during prompt initialization to penalize excessively large initial values, thereby achieving more stable and generalizable prompts:

$$\mathcal{L}_{L2} = \|\mathbf{p}\|_2 = \sqrt{\sum_{i=1}^{L_p} \boldsymbol{p}_n^2} \tag{13}$$

Here $\boldsymbol{p}_n$ is the element of prompt, $L_p$ is the dimension of the prompt. The total lossfunction is:

$$\mathcal{L} = \mathcal{L}_{CE} + \lambda \mathcal{L}_{L2} \tag{14}$$

where $\lambda$ is the regularization coefficient used to control the weight of the $L2$ regularization term.

After training of $i$th task $\mathcal{D}_i$, mainstream prompt-based methods learn the keys based on Eq. (5), which introduces additional training and memory overhead. Unlike training the key-value pairs after completing the training, we directly compute the prototype of the current task under prompt $\boldsymbol{p}_i$ using Eqs. (7) and (8), and set the prototype as the weights of the classifier $\phi(\boldsymbol{x})$ for inference. In this way, the representative features of task $\mathcal{D}_i$ under prompt $\boldsymbol{p}_i$ are extracted and stored within the weights of classifier. During prediction, by calculating the similarity between the features and the prototypes using Eq. (11), we can determine the probability that the data belongs to each class.

**Algorithm 1** ProP at training time.

**Input:** Incremental datasets: $\{\mathcal{D}_1, \mathcal{D}_2, \ldots, \mathcal{D}_T\}$, Pre-trained embedding: $f(\theta)$;
**Output:** Finetuned model $f(\theta, \boldsymbol{P})$

1: **for** $t = 1, 2, \ldots, T$ **do**
2: Get the incremental training task $\mathcal{D}_t$;
3: Initialize a new prompt $\boldsymbol{p}_t$;
4: Optimize the model $f(\theta, \boldsymbol{p}_t)$ via Eq. (14);
5: Extract the prototypes $\boldsymbol{C}_t$ of $\mathcal{D}_t$ for prompt $\boldsymbol{p}_t$ via Eqs. (7)–(9);
6: Replace the weight of classifier $\phi(x)$ with $\boldsymbol{C}_t$;
7: **end for**
8: **return** the updated model $f(\theta, \boldsymbol{P})$;

### 3.3. Pseudo code

We summarize the training and testing processes of ProP in Algorithms 1 and 2. As illustrated in Algorithm 1, we initialize a task-specific

**Algorithm 2** ProP at testing time.

**Input:** Incremental datasets: $\mathcal{D}_t$, Finetuned model: $f(\theta, \boldsymbol{P})$;
**Output:** Prediction

1: **for** $i = 1, 2, \ldots, t$ **do**
2: Abstract feature $\boldsymbol{H}_i$ via Eq. (10);
3: Extract prototype $\boldsymbol{C}_i$ form classifier $\phi(x)$;
4: Calculate similarity via Eq. (11);
5: **end for**
6: **return** prediction;

prompt $\boldsymbol{p}_t$ for task $\mathcal{D}_t$ (line 3) and train it using cross-entropy loss and L2 loss (line 4). Subsequently, we extract the task-specific prototype $C_t$ for task $\mathcal{D}_t$ (line 5) and replace the weights of classifier with it (line 6). Notably, by using the prototype as the weight of classifier, the classifier in the training stage will be dropped out.

At the testing stage in Algorithm 2, we already get trained task-specific prompt $\boldsymbol{P}$. Therefore, we can abstract feature $\boldsymbol{H}_i$ of input with task-specific prompt $\boldsymbol{p}_i$ (line 2). We also can directly extract the task-specific prototype $\boldsymbol{C}_i$ from the classifier (line 3). By calculating the similarity between the feature and prototype, we can determine the class the test data belongs to.

### 3.4. Computational complexity

To rigorously evaluate the efficiency of ProP, we conduct a detailed analysis of its computational complexity and compare it with mainstream key-value based prompting methods. The inference process of ProP mainly consists of two computational stages. First, it computes the similarity between the input feature and all task prototypes $\boldsymbol{C}_i$, which incurs a complexity of $\mathcal{O}(T \times L_p \times D)$. Second, it generates a task-specific feature representation $H_{p_i}$ for each task by concatenating the input with the corresponding task-specific prompt $p_i$ and performing a forward pass through the model. This stage has a complexity of $\mathcal{O}(L \times (L_h + L_p)^2 \times D)$. Therefore, the total computational cost of ProP is the sum of these two components, where $L$ denotes the number of Transformer layers. In contrast, key-value based methods such as L2P first compute a query vector and match it with $P$ keys in the prompt pool, resulting in a complexity of $\mathcal{O}(P \times L_p \times D)$. The top-$k$ retrieved prompts are then used for a single forward propagation, whose complexity is $\mathcal{O}(L \times (L_h + k \times L_p)^2 \times D)$.

Although the overall complexity of ProP scales linearly with the number of tasks $T$, its computations are highly parallelizable. More importantly, by removing the key-value matching step, ProP avoids potential matching errors and allocates the saved resources directly to enhancing the final classification accuracy. As demonstrated in Fig. 2 and Table 1, ProP outperforms other methods across multiple benchmarks, validating the effectiveness of this computational trade-off.

## 4. Experiments

In this section, we first present the implementation details of ProP in Section 4.1. Then we compare our approach with state-of-the-art methods across multiple benchmark datasets to demonstrate its incremental learning capabilities in Section 4.2. Additionally, we conduct ablation experiments to validate the robustness of our method in Section 4.3. Finally, we present visualization and seed average in Section 4.4.

### 4.1. Implementation details

**Benchmark.** We evaluate model performance on several benchmark datasets, including CIFAR-100 (Krizhevsky & Hinton, 2009), CUB-200 (Wah et al., 2011), ImageNet-A (Hendrycks et al., 2021a), ImageNet-R (Hendrycks et al., 2021b), ObjectNet (Barbu et al., 2019), OmniBenchmark (Zhang et al., 2022), and VTAB (Zhai et al., 2019). Since the majority of pre-trained models are trained on ImageNet, datasets like ImageNet-A and ObjectNet contain challenging samples that are difficult for pre-trained modes to handle. Additionally, OmniBenchmark and VTAB encompass diverse classes from multiple complex domains. Specifically, CIFAR-100 includes 100 classes, CUB-200, ImageNet-R each include 200 classes. Following Zhou et al. (2024a), we sample 200 classes from ObjectNet and ImageNet-A and 300 classes from Omnibenchmark to construct the CIL benchmark, and we also sample 5 datasets from VTAB, each dataset containing 10 classes.

In our benchmark settings, following Rebuffi et al. (2017), we adopt the notation 'Init $i$ Inc $j$', where $i$ denotes the number of classes in the first stage and $j$ indicates the number of classes in each incremental stage. The notation 'Init $i$, Inc $j$' signifies that the first dataset comprises $i$ classes, while each subsequent dataset contains $j$ classes.

**Baseline** We begin by comparing ProP with state-of-the-art prompt-based CIL methods, including L2P (Wang et al., 2022d), DualPrompt, Wang et al. (2022c) CODA-Prompt (Smith et al., 2023), and APER (Zhou et al., 2024a). Additionally, we evaluate it against conventional CIL methods, such as LwF (Li & Hoiem, 2017), SDC (Yu et al., 2020), iCarl (Rebuffi et al., 2017), and DER (Yan et al., 2021), which also leverage the same pre-trained model for initialization. We also report baseline results obtained by fully finetuning the pre-trained model, referred to as Finetune. All experiments are conducted using the same pre-trained model to ensure consistency.

**Training details** Following Wang et al. (2022d), Zhou et al. (2024a), all experiments are conducted using ViT-B/16 pre-trained on ImageNet-21k, denoted as "ViT-B/16-IN21K," and "ViT-B/16-IN1K," which is further finetuned on ImageNet-1K. During finetuning, we use Stochastic Gradient Descent (SGD) as the optimizer, with an initial learning rate of 0.03 and a cosine decay schedule. The weight decay is set to 0.0005, and each prompt has a length of $L_p = 5$. Furthermore, the training process utilizes a batch size of 16, runs for 20 epochs per session, and employs an L2 regularization coefficient ($\lambda$) of 0.1.

**Network architectures in the ProP.** We developed ProP, which leverages ViT (Dosovitskiy, 2020) and task-specific prompt-prototype to perform class-incremental learning tasks. In ProP, the encoder comprises both a pre-trained model and a finetuned model. We utilize these two models to compute the features of input separately and then directly concatenate the features together, as illustrated in the Eq. (10). By concatenating pre-trained model, we maximally retain the feature extraction capacity. Additionally, we employ a fully connected layer as the classifier, where the weights of the classifier correspond to the class prototypes. At the training stage, we first randomly initial the classifier. After training, we calculate the prototype vectors of current task and replace the weights of the classifier with prototypes. Similarly to feature extraction, we compute the prototypes from both the pre-trained model and the finetuned model separately and then directly concatenate them together, as shown in the Eq. (9).

**Evaluation Metric**: In accordance with the benchmark protocol, we follow Rebuffi et al. (2017), using Average accuracy (**Avg**) and Last accuracy (**Last**) to measure the performance of model. Specifically, **Last** denotes the Top-1 accuracy of every task, and **Avg** is the average value of **Last** of all tasks. Mathematically, for the $t$th task, Average accuracy is calculated as follows: $\textbf{Avg}_t = \frac{1}{t}\sum_{i=1}^{t}\textbf{Last}_i$.

### 4.2. Comparison with other methods

We compared our method with traditional approaches and prompt-based methods across a wide range of datasets, with the results presented in Table 1. It can be observed that our method demonstrates the best performance in the vast majority of cases, particularly on datasets where ViT performs poorly, such as ImageNet-A and ImageNet-R, where we observed an average improvement of over 5%. This indicates that, by binding the prompt with prototypes sharing same task id, the model can learn effective feature representations that the pre-trained model did not capture on the pre-trained datasets, significantly mitigating the domain gap between the pre-trained datasets and the continual learning downstream tasks. Additionally, we also clearly observe that prompt-based methods generally outperform traditional continual learning methods, suggesting that prompts offer a more effective means of fine-tuning pre-trained models.

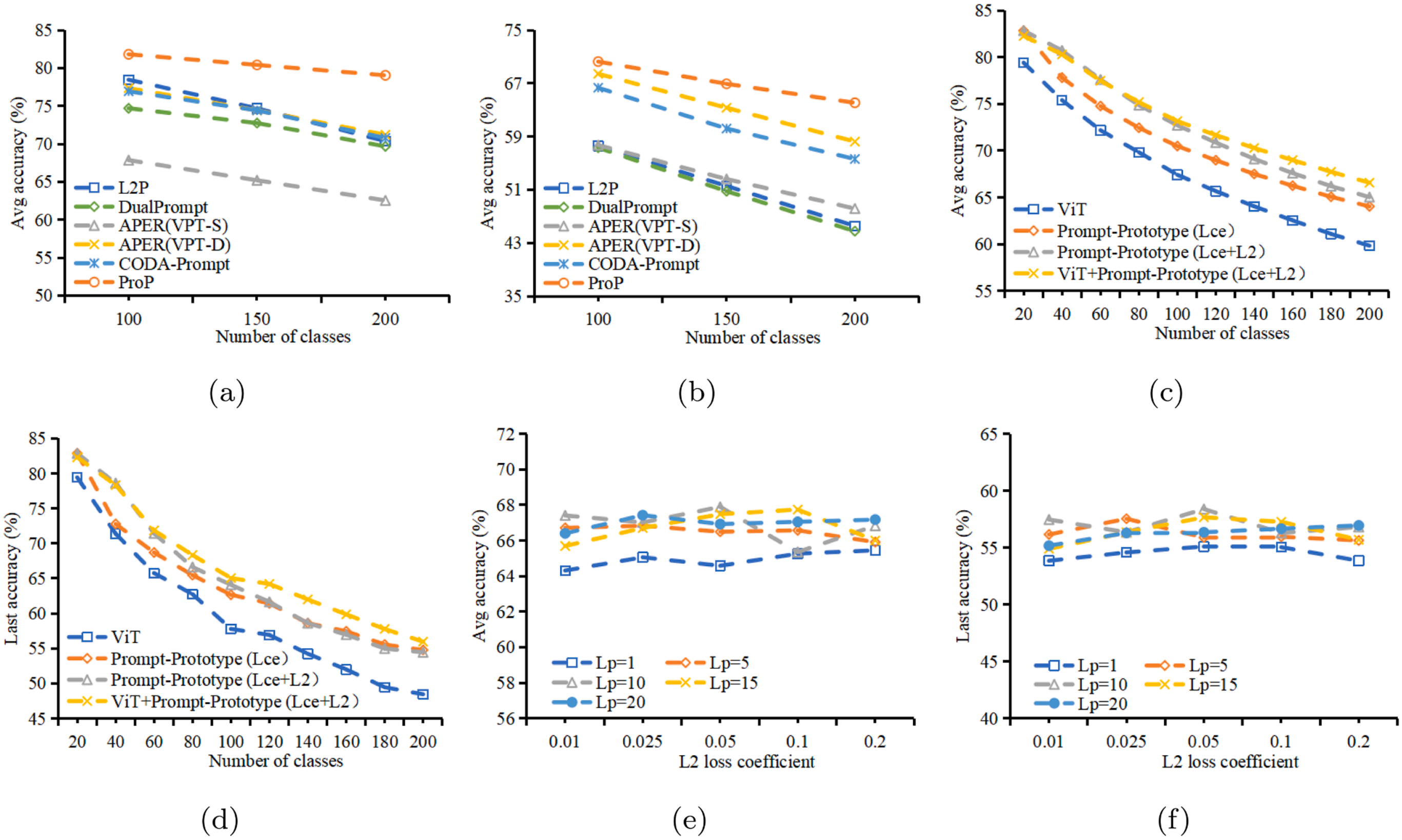

**Fig. 4.** Results on different settings. Here, (a)(b) presents the average accuracy on large initial classes in ImageNet-R(a) and ImageNet-A(b) Init100 Inc50. All experiments are conducted based on **ViT-B/16-IN21K**. (c)(d) study the impact of components proposed in ProP, where "ViT" represents using the pre-trained **ViT-B/16-IN21K** for CIL without finetuning, while "Prompt-Prototype" indicates **ViT-B/16-IN21K** finetuned with task-specific prompt-prototype. "Lce" refers to training with only cross-entropy loss $\mathcal{L}_{CE}$ and "Lce + L2" denotes training with both cross-entropy loss $\mathcal{L}_{CE}$ and $L2$ loss $\mathcal{L}_{L2}$. (e)(f) show the influence of L2 loss coefficient and prompt length $L_p$ on model performance. Experiments are conducted on ImageNet-A Init20 Inc20 with **ViT-B/16-IN21K** as backbone.

The choice of backbone architecture and incremental learning configuration has a substantial impact on the overall performance of continual learning models. We adopt ViT-B/16-IN1K as the base model and compare ProP with other representative approaches under different incremental learning settings, and the results are illustrated in Fig. 3. Specifically, ViT-B/16-IN21K is pre-trained on the ImageNet-21K dataset, while ViT-B/16-IN1K is further finetuned on ImageNet-1K following the ImageNet-21K pre-training stage. This additional finetuning alters the model's attention patterns, leading to noticeable performance variations across tasks. We evaluate ProP using the ViT-B/16-IN1K backbone under multiple incremental learning settings to examine its robustness. As shown in Fig. 3, ProP consistently achieves superior performance across different initial class configurations, indicating that its effectiveness does not depend on a specific backbone and that it exhibits strong robustness. Moreover, slight performance fluctuations can be observed when using ViT-B/16-IN1K as the base model. For instance, as shown in Table 1, when ViT-B/16-IN21K serves as the backbone, ProP surpasses other methods by more than 5% on the ImageNet-A and ImageNet-R datasets, whereas this advantage becomes less pronounced when using ViT-B/16-IN1K. These observations suggest that both the choice of backbone and the incremental learning configuration significantly influence the model's ability to mitigate catastrophic forgetting.

**Table 2**
Average and last performance comparison with rehearsal methods with **ViT-B/16-IN21K** as the backbone. "Exemplars" presents number of sample stored. The best performance is shown in bold.

| | | CIFAR Init5 Inc5 | | IN-R Init20 Inc20 | |
|---|---|---|---|---|---|
| Method | Exemplars | Avg | Last | Avg | Last |
| iCaRL | 20/class | 72.42 | 60.67 | 82.46 | 73.87 |
| DER | 20/class | 80.48 | 74.32 | 86.04 | 77.93 |
| FOSTER | 20/class | 81.34 | 74.48 | 89.87 | 84.91 |
| MEMO | 20/class | 74.80 | 66.62 | 84.08 | 75.79 |
| ProP | 0 | **91.84** | **85.99** | **91.84** | **85.99** |

Furthermore, in addition to the settings where the initial and incremental classes are equal, as illustrated in the figures and tables, we also evaluated the model's performance when the number of initial and incremental classes differed, and the results are shown in Fig. 4(a), (b). From the figure, it is evident that compared to other CIL methods, ProP consistently exhibits a smaller decrease in accuracy during the learning process, indicating its superior anti-forgetting capabilities. Additionally, we compared ProP with replay-based continual learning methods, which retain previous data for rehearsal and typically demonstrate good performance. The replay-based methods employing the same pre-trained model initialization, and the results are presented in Table 2. Surprisingly, ProP achieved the highest accuracy without the need to store examples, further demonstrating its outstanding anti-forgetting performance.

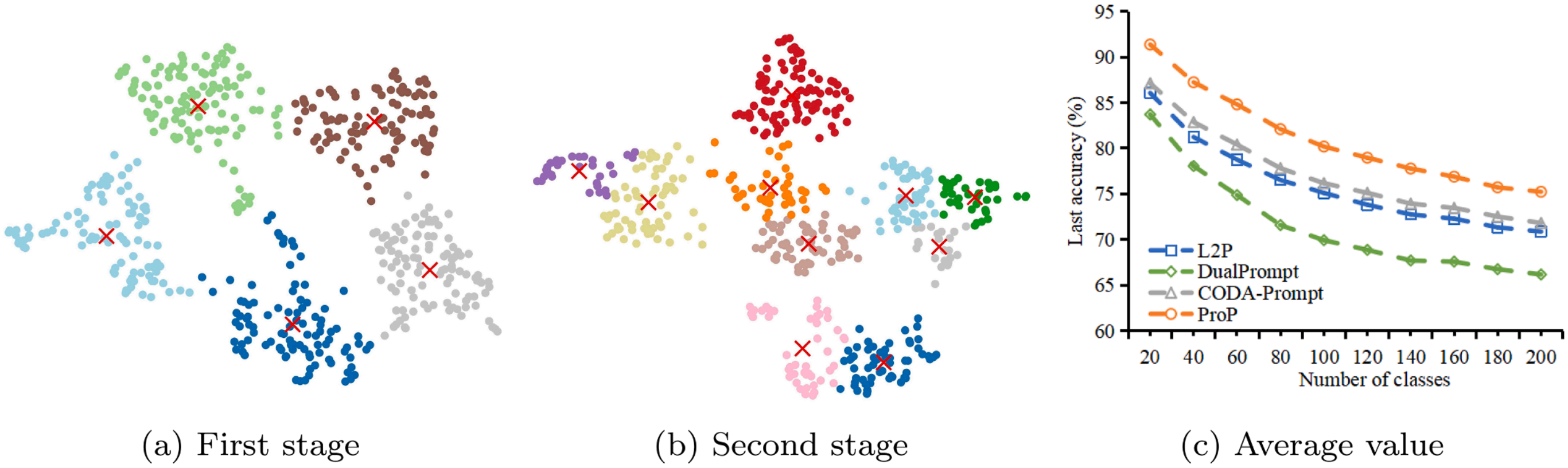

**Fig. 5.** Visualization and average value. (a) and (b) are visualization of the feature and class prototypes at first and second stage on CIFAR Init5 Inc5. Test data is represented by dots while class prototypes are present by "×", (c) is average value of last accuracy under different seeds based on ImageNet-R Init20 Inc20 with **ViT-B/16-IN21K** as backbone.

**Table 3**
Performance under different feature fusion methods with **ViT-B/16-IN21K** as backbone. 'Concatenate' denotes the method used in this work. The best results are shown in bold.

| | CIFAR Init5 Inc5 | | CUB Init10 Inc10 | | IN-A Init20 Inc20 | |
|---|---|---|---|---|---|---|
| Method | Avg | Last | Avg | Last | Avg | Last |
| Sum | 91.81 | 85.82 | 91.05 | 84.86 | 66.23 | 55.37 |
| Pooling | 91.26 | 84.99 | 91.19 | **84.99** | 65.19 | 54.25 |
| Average | 91.83 | 85.44 | 89.26 | 82.06 | 66.19 | 55.23 |
| Concatenate | **91.84** | **85.99** | **91.20** | 84.82 | **66.57** | **55.96** |

### 4.3. Ablation study

We analyze the proposed modules in this section, with results in Fig. 4(c)(d). "ViT" refers to using ViT-B/16-IN21K without finetuning, while "Prompt-Prototype" finetunes it with task-specific prompt-prototypes. "ViT + Prompt-Prototype" concatenates features from both the pre-trained and finetuned models, as in Eq. (10). "Lce" uses only cross-entropy loss $\mathcal{L}_{CE}$, whereas "Lce + L2" adds L2 loss $\mathcal{L}_{L2}$. Experiments follow ImageNet-A Init20 Inc20 settings from Section 4.1. Fig. 4(c), (d) shows that task-specific prompts significantly boost performance over the pre-trained model, demonstrating the benefits of binding prompts with prototypes of the same task. Adding $\mathcal{L}_{L2}$ further improves finetuning by stabilizing prompt initialization. Finally, concatenating pre-trained and finetuned features maximizes knowledge retention, enhancing overall performance.

To investigate the trade-off between generalization and robustness, we analyze different strategies for fusing prototype features extracted from the finetuned model and the pre-trained model. The finetuned model effectively captures task-specific representations from the downstream dataset but inevitably forgets previously learned knowledge. In contrast, the pre-trained model retains prior knowledge yet lacks sufficient adaptability to the new task. To better integrate the complementary strengths of both, we experiment with several prototype feature fusion methods, as summarized in Table 3. Specifically, 'Sum' denotes element-wise addition of features, 'Pooling' refers to max-pooling across feature dimensions, 'Average' represents feature-wise mean fusion, and 'Concatenate' indicates direct feature concatenation. As shown in Table 3, the concatenation-based fusion consistently outperforms the other approaches across all evaluation benchmarks. This result suggests that concatenation more effectively balances generalization and robustness by preserving diverse information from both sources. Therefore, in this work, we adopt the concatenation strategy for prototype feature fusion.

We further investigated the impact of the L2 loss coefficient $\lambda$ and the prompt length $L_p$ on the model. Specifically, the L2 coefficients $\lambda$ evaluated are {0.01, 0.05, 0.025, 0.1, 0.2}, and the prompt lengths $L_p$ tested are {1, 5, 10, 15, 20}. The experiments were conducted using ViT-B/16-IN21K on the ImageNet-A Init20 Inc20, and the results are presented in Fig. 4(e), (f). From the figure, we can observe that the inclusion of L2 loss effectively enhances the performance of model, indicating that penalizing excessively large initial values facilitates a more effective prompt learning process.

Furthermore, the model does not exhibit significant sensitivity to variations in the L2 loss coefficient, demonstrating good robustness to L2 loss. Additionally, we found that results of prompt length $L_p = 1$ are in noticeably poorer performance. However, when the prompt length is increased to 5 or larger, no specific length shows a clear advantage. This suggests that there is no apparent positive correlation between prompt length and model performance, therefore simply increasing the prompt length does not continuously improve the performance of model. Therefore, in this study, we selected $\lambda = 0.1$ and $L_p = 5$.

### 4.4. Visualization and seed average

Additionally, we employed t-SNE (Maaten & Hinton, 2008) to visualize the first two stages of the incremental learning process of ProP. Fig. 5(a) represents the visualization of the first stage, while Fig. 5(b) depicts the second stage. From the figures, we can observe that ProP effectively categorizes the data into their respective classes during the learning process, demonstrating that the model has learned robust feature representations for each class and efficiently divided different classes based on these. In the second stage, the model still can recognize it well. Furthermore, we can also observe that the class prototypes are generally positioned at the centres of their corresponding data clusters, indicating that the class prototypes accurately represent the distinctive features of each class.

During the training stage of ProP, we follow the approach described in Fig. 5(c) to randomly shuffle the order of all datasets using a fixed random seed of 1993, ensuring that the training sequence remains unbiased. To further ensure a fair and robust comparison with other CIL methods, we also conduct experiments using multiple random seeds (*i.e.*, {1994, 1995, 1996, 1997}) and report the average final accuracy, as illustrated in Fig. 5(c). As we can see, the results in Fig. 5(c) clearly show that ProP consistently outperforms other CIL methods, highlighting not only its superior performance but also its robust stability across different random initialization scenarios.

## 5. Conclusion

In this paper, we propose ProP, a prompt-based continual learning framework that does not require key-value pair. By binding task-specific prompt with prototype, we eliminate the inter-task information interference and scalability limitations introduced by key-value pair. Additionally, we incorporate an $L2$ loss during the prompt initialization process to penalize excessively large initial values, resulting in more stable and generalizable prompts. Experiments conducted on numerous datasets demonstrate the effectiveness of ProP. Unlike mainstream prompt-based approaches, our method removes the dependence on key-value pairing, offering a new perspective for future research.

## Acknowledgments

This work was supported in part by the National Natural Science Foundation of China (NSFC) under Grant (No. 62476035, and U24B20140), and the Young Elite Scientists Sponsorship Program by CAST under Grant 2024QNRC001.

## CRediT authorship contribution statement

**Haihua Luo:** Writing – review & editing, Writing – original draft, Visualization, Validation, Project administration, Methodology, Investigation, Formal analysis, Data curation, Conceptualization; **Xuming Ran:** Writing – review & editing, Writing – original draft, Validation, Supervision, Project administration, Methodology, Formal analysis, Data curation, Conceptualization; **Zhengji Li:** Writing – original draft, Visualization, Validation, Software, Resources, Project administration; **Huiyan Xue:** Writing – original draft, Visualization, Validation, Investigation; **Tingting Jiang:** Writing – original draft, Validation, Software, Investigation; **Jiangrong Shen:** Writing – review & editing, Supervision, Resources, Project administration, Methodology; **Tommi Kärkkäinen:** Writing – review & editing, Supervision, Resources, Project administration, Funding acquisition; **Qi Xu:** Writing – review & editing, Supervision, Resources, Project administration, Methodology, Formal analysis, Conceptualization; **Fengyu Cong:** Writing – review & editing, Supervision, Resources, Funding acquisition.

## Declaration of competing interest

The authors declare that they have no known competing financial interests or personal relationships that could have appeared to influence the work reported in this paper.